\documentclass[runningheads]{llncs}
\usepackage{graphicx}

\usepackage{tikz}
\usepackage{comment}
\usepackage{amsmath,amssymb} 
\usepackage{color}
\usepackage{multirow}
\usepackage{subfig,graphicx}

\usepackage{booktabs}
\usepackage[hidelinks,colorlinks, linkcolor=cyan, anchorcolor=cyan, citecolor=cyan]{hyperref}
\renewcommand{\thefigure}{S\arabic{figure}}

\renewcommand{\thetable}{S\arabic{table}}

\begin{document}
\pagestyle{headings}
\mainmatter
\def\ECCVSubNumber{4958}  

\title{Outpainting by Queries} 


\titlerunning{Outpainting by Queries}
%
\author{Kai Yao\inst{1,2} \and
Penglei Gao\inst{1,2} \and
Xi Yang\inst{2} \and
Kaizhu Huang\inst{3} \and
Jie Sun\inst{2} \and
Rui Zhang\inst{2}}
\authorrunning{K. Yao et al.}
%
\institute{University of Liverpool, Liverpool L69 7ZX, U.K.
\email{\{Kai.Yao,P.Gao6\}@liverpool.ac.uk} \and
Xi'an Jiaotong Liverpool University, Suzhou, 215123, China \and
Duke Kunshan University, Kunshan 215316, China \\
}
\maketitle

\renewcommand{\thefootnote}{}
\footnotetext{K. Yao and P. Gao --- Equal Contribution.}

\begin{abstract}
Image outpainting, which is well studied with Convolution Neural Network (CNN) based framework, has recently drawn
more attention in computer vision. However, CNNs rely on inherent inductive biases to achieve effective sample learning, which may degrade the performance ceiling.
In this paper, motivated by the flexible self-attention mechanism with minimal inductive biases in transformer architecture, we reframe the generalised image outpainting problem as a patch-wise sequence-to-sequence autoregression problem, enabling query-based image outpainting.
Specifically, we propose a novel hybrid vision-transformer-based encoder-decoder framework, named \textbf{Query} \textbf{O}utpainting \textbf{TR}ansformer (\textbf{QueryOTR}), for extrapolating visual context all-side around a given image.
Patch-wise mode's global modeling capacity allows us to extrapolate images from the attention mechanism's query standpoint.
A novel Query Expansion Module (QEM) is designed to integrate information from the predicted queries based on the encoder's output, hence accelerating the convergence of the pure transformer even with a relatively small dataset.
To further enhance connectivity between each patch, the proposed Patch Smoothing Module (PSM) re-allocates and averages the overlapped regions, thus providing seamless predicted images. We experimentally show that QueryOTR could generate visually appealing results smoothly and realistically against the state-of-the-art image outpainting approaches.
Code is available at \url{https://github.com/Kaiseem/QueryOTR}.
\end{abstract}


\section{Introduction}
Image outpainting, usually known as image extrapolation, is a challenging task that requires extending image boundaries by generating new visually harmonious contents with semantically meaningful structure from a restricted input image.
It could be widely applied in the real world to enrich humans' social lives based on limited visual content, such as automatic creative image, virtual reality, and video generation \cite{ma2021boosting}.
Different from image inpainting \cite{bertalmio2000image,barnes2009patchmatch,pathak2016context,yu2018generative}, which could take advantage of  visual contexts surrounding an inpainting area, generalised image outpainting should extrapolate the unknown regions in all directions around the sub-image.
As the unknown pixels farther from the image borders are less constrained, they have a greater chance of accumulating expanded-errors or generating repetitive patterns than those closer to the borders.
Consequently, the challenges of this task include: (a) determining where the missing features should be located relative to the output's spatial locations for both nearby and faraway features; (b) guaranteeing that the extrapolated image has a realistic appearance with reasonable content and a consistent structural layout with the conditional sub-image; and (c) the borders between extrapolated regions and the original sub-image should be smooth and seamless.

Convolutional architectures have been proven successful for computer vision tasks nowadays. Existing image outpainting methods utilize kinds of variants of CNN-based methods to conduct image extrapolation. CNNs rely on inherent inductive biases to achieve effective sample learning, which may degrade the performance ceiling. Although the existing CNN-based outpainting methods achieve solid performance \cite{IOHGan,NSIPO,SRN,kim2021painting,ma2021boosting}, they still suffer from blunt structures and abrupt colours when extrapolating the unknown regions of the images. The potential reason might be that the inductive biases of convolution in such CNN-based architectures are hard-coded in the form of two strong constraints on the weights: locality and weight sharing \cite{d2021convit}. These constraints may degrade the model's ability to represent global features and capture long-range dependencies.


\begin{figure}[t]
\centering
\includegraphics[width=\columnwidth]{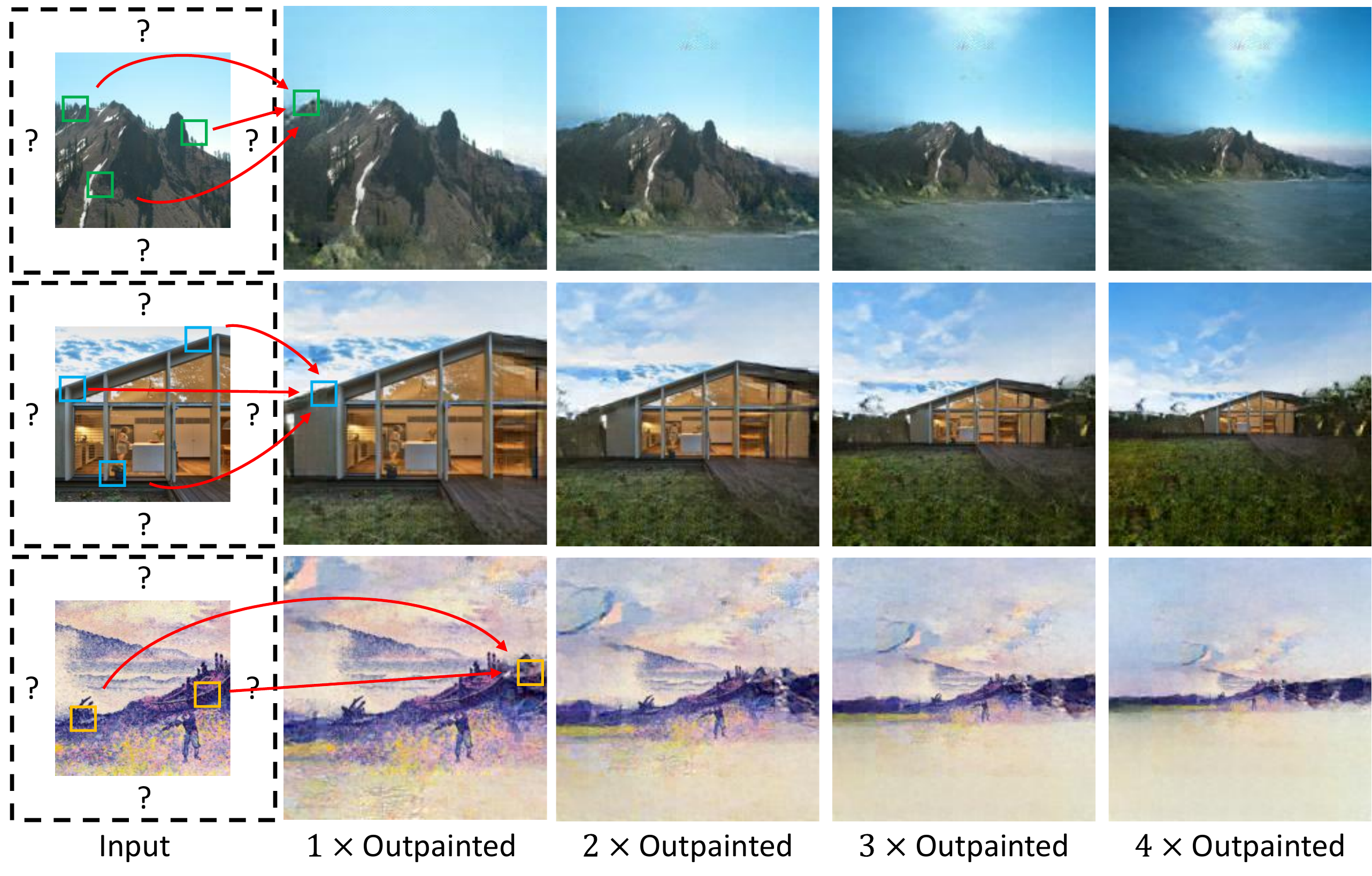}
\caption{Demonstration of recursive outpainting by our QueryOTR. Our method generates a sequence of extrapolated image patches by querying the sequence of input image patches, enabling a remarkable perceptual consistency.}
\label{fig:vis1}
\end{figure}

Transformer architectures have competitive performance in areas such as image and video recognition. The transformer dispenses with the convolutional inductive bias by performing self-attention across embeddings of patches of pixels, which breaks through the limitation of capturing long-range dependencies. However, in the pure transformer, the model converges very slowly with a relatively small dataset~\cite{d2021convit}.
On the ImageNet benchmark, Dosovitskiy et al.~\cite{vit} developed the Vision Transformer (ViT) interpreting a picture as a sequence of tokens, which can achieve comparable image classification accuracy while requiring less computational budgets.
ViT relies on globally-contextualized representation, in which each patch is attended to all patches of the same image, as opposed to local-connectivity in CNNs. ViT and its variants have shown promising superiority in modeling non-local contextual relationships as well as good efficiency and scalability, though they are still in their infancy. In light of the global interaction and the generation of distant features with conditional sub-image, these benefits could enhance image extrapolation in a beneficial fashion.

\begin{figure}[!t]
\centering
\begin{minipage}[]{.52\linewidth}
\centering
\subfloat[][Autoregression w/ and w/o QEM]{\label{vis:a}
\includegraphics[width=\columnwidth]{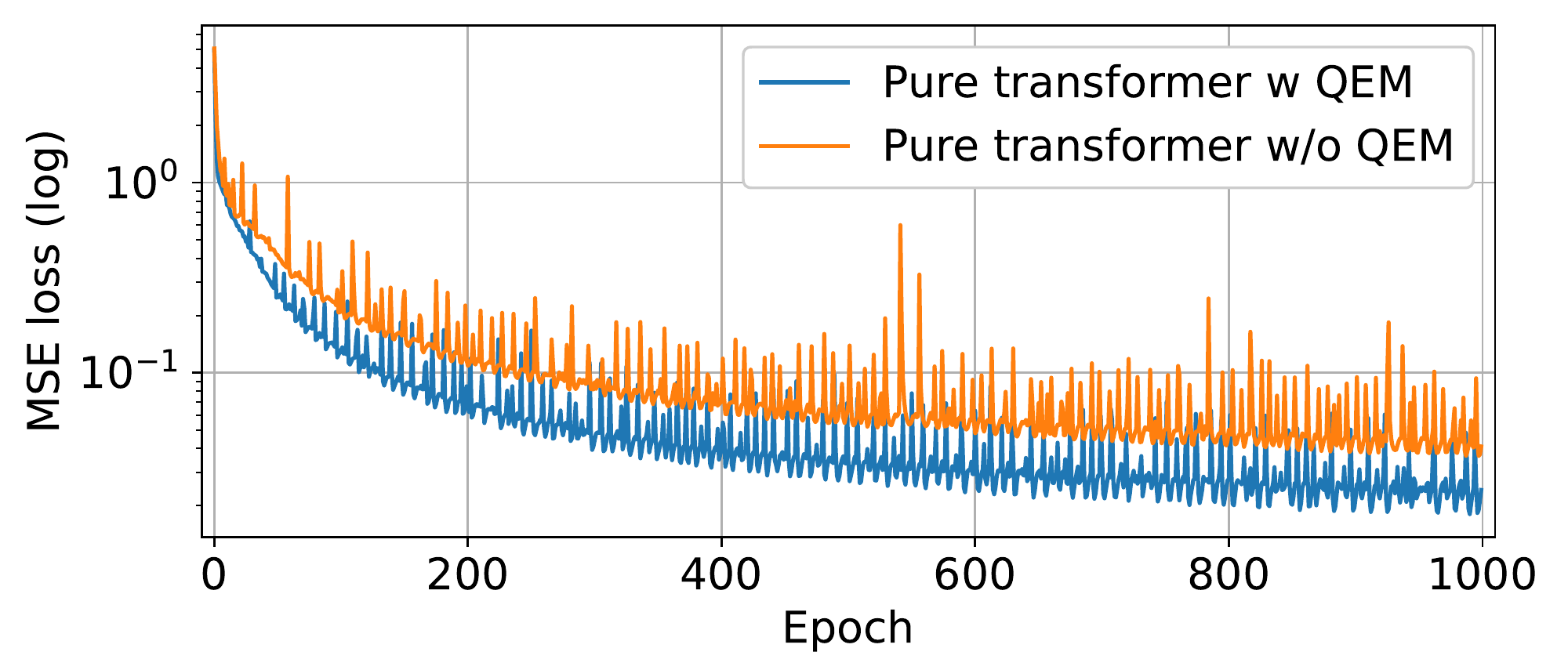}}
\end{minipage}\hfill
\begin{minipage}[]{.22\linewidth}
\centering
\subfloat[][w/o PSM]{\label{vis:b}
\includegraphics[width=\columnwidth]{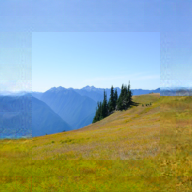}}
\end{minipage}\hfill
\begin{minipage}[]{.22\linewidth}
\centering
\subfloat[][w/ PSM]{\label{vis:c}
\includegraphics[width=\columnwidth]{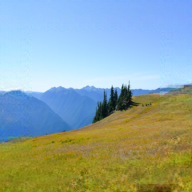}}
\end{minipage}\hfill
\caption{(a) Training a pure transformer encoder-decoder with and without QEM to regress unseen image patches. QEM significantly speeds up the convergence (about 3.3 times faster than that without QEM: w/ QEM at 300 epoch v.s. w/o QEM at 1,000 epoch). (b) QueryOTR without PSM. (c) QueryOTR with PSM.}
\label{vis}
\end{figure}

To better cope with image long-range dependencies and spatial relationships between predicted regions and conditional sub-images, we reconsider the outpainting problem as a patch-wise sequence-to-sequence autoregression problem inspired by the original transformer~\cite{aiayn} in natural language processing.
We develop a novel hybrid query-based encoder-decoder transformer framework, named \textbf{Query} \textbf{O}utpainting \textbf{TR}ansformer (\textbf{QueryOTR}), to extrapolate visual context all-side around a given image taking advantages of both ViT~\cite{vit} and pure transformer~\cite{aiayn} in the image outpainting task, as shown in \autoref{fig:vis1}.
Specifically, we design two special modules, Query Expansion Module (QEM) and Patch Smoothing Module (PSM), to conduct feature forecasting from the perspective of the query in the attention mechanism. 
In contrast to the query learning in pure transformer, our designed query in QEM is predicted by the stacked CNN-based blocks based on the output of the transformer encoder. The predicted query is easy to learn and has better flexibility by drawing on the advantages of CNNs' inductive biases to accelerate query prediction converge in pure transformer for approximately three times faster than that without QEM in training, which is shown in \autoref{vis}(a).
The developed PSM re-allocates the predicted patches around the conditional sub-image and averages the overlapping parts to make the generated image smoothly and seamlessly. Also, PSM contributes to alleviate the problem of checkerboard artifact caused by the independent procession among the output image patches. In this way, the model could focus more on the connections between each patch and enhance the representing ability as shown in \autoref{vis}(b) and (c). 
Our \textbf{QueryOTR} is the first hybrid transformer as a sequence-to-sequence modeling, which is able to extend image borders seamlessly and generate unseen images smoothly and realistically.

The main contributions of this work are three-fold:
\begin{itemize}
\item We rephrase the image outpainting problem as a patch-wise sequence-to-sequence autoregression problem and develop a novel hybrid transformer encoder-decoder framework, named \textbf{QueryOTR}, for query-based prediction of extrapolated images, and minimization of degradation from the inductive biases in CNN-structures.
\item We propose \textit{Query Expansion Module} and \textit{Patch Smoothing Module} to solve the slow convergence problem in pure transformers and to generate realistic extrapolated images smoothly and seamlessly.
\item Experimental results show that the proposed method achieves state-of-the-art one-step and multi-step outpainting performance as compared to recent image outpainting methods.
\end{itemize}

\section{Related Work}
\label{sec:Rela}
\subsection{Image Outpainting}
Generative Adversarial Networks (GANs) \cite{goodfellow2014generative} have been widely applied in many research fields, such as image super-resolution, image synthesis, and image denoising \cite{gu2020image,ledig2017photo,brock2018large,lsgan,gulrajani2017improved}. Efforts have been made for image generation with  GAN under certain conditions. Image extrapolation aims to generate the surrounding regions from the visual content, which can be considered as an image-conditioned generation task \cite{guo2020spiral}.
Sabini and Rusak~\cite{sabini2018painting} brought the image outpainting task into public attention with a deep neural network framework inspired by the image inpainting methods. This effort focused on enhancing the quality of generated images smoothly by using GANs and the post-processing methods to perform horizontal outpainting.
Van et al.~\cite{IOHGan} designed a CNN-based encoder-to-decoder framework by using GAN for image outpainting.
Wang et al.~\cite{SRN} proposed a Semantic Regeneration Network to directly learn the semantic features from the conditional sub-image.
Han et al.~\cite{lin2021edge} developed a 3-stage deep learning model with an edge-guided generative network to produce semantically consistent output from a small image input.
Although these methods avoid the bias in the general padding and up-sampling pattern, they still suffer from blunt structures and abrupt colours issues, which tend to ignore the spatial and semantic consistency.
To tackle these issues, Yang et al.~\cite{NSIPO} proposed a Recurrent Content Transfer (RCT) block for temporal content prediction with Long Short Term Memory (LSTM) networks as the bottleneck. To increase the contextual information, Lu et al.~\cite{lu2021bridging} and Kim et. al.~\cite{kim2021painting} rearranged the boundary region by switching the outer area of the image into its inner area. 
These latest models are based on convolutional neural networks. As global information is not well captured, they all have limitations in explicitly modelling long-range dependency.

\subsection{Transformer}
Recently, transformer has attracted much attention in computer vision. Transformer was first proposed to solve NLP tasks by replacing the traditional CNN and Recurrent Neural Network (RNN) structures \cite{aiayn}. The Self-Attention mechanism helps the model learn the global representation from the input which could improve the performance for basic visual feature extraction~\cite{aiayn}.
Jacob et al.~\cite{bert} introduced a very deep network to pretrain deep bidirectional representations from unlabeled text by jointly conditioning on both left and right context in all layers. It can be fine-tuned with just one additional output layer for better performance.
ViT \cite{vit} is a convolution-free Transformer that conducts image classification over a sequence of image patches. The superiority of the Transformer architecture is presented in ViT fully utilizing the advantage of pretraining on large-scale datasets compared with the CNN-based methods.
Many ViT-based variants also demonstrated the success in computer vision tasks \cite{DeepViT,PiT,LeViT}, such as object detection \cite{detr}, video recognition \cite{vivit}, and image synthesis \cite{vitgan}.
Moreover, Liu et al.~\cite{swin} proposed Swin Transformer to extend vision tasks for object detection and semantic segmentation.
Gao et al.~\cite{UTransformer} designed a transformer-based framework for image outpainting with an encoder-decoder architecture. They used Swin Transformer which involved shifted window attention to bridge the windows of the preceding layer, which significantly enhanced modelling power as well as achieved lower latency.


\section{Methodology}
\begin{figure*}[!t]
\centering
\includegraphics[width=1\columnwidth]{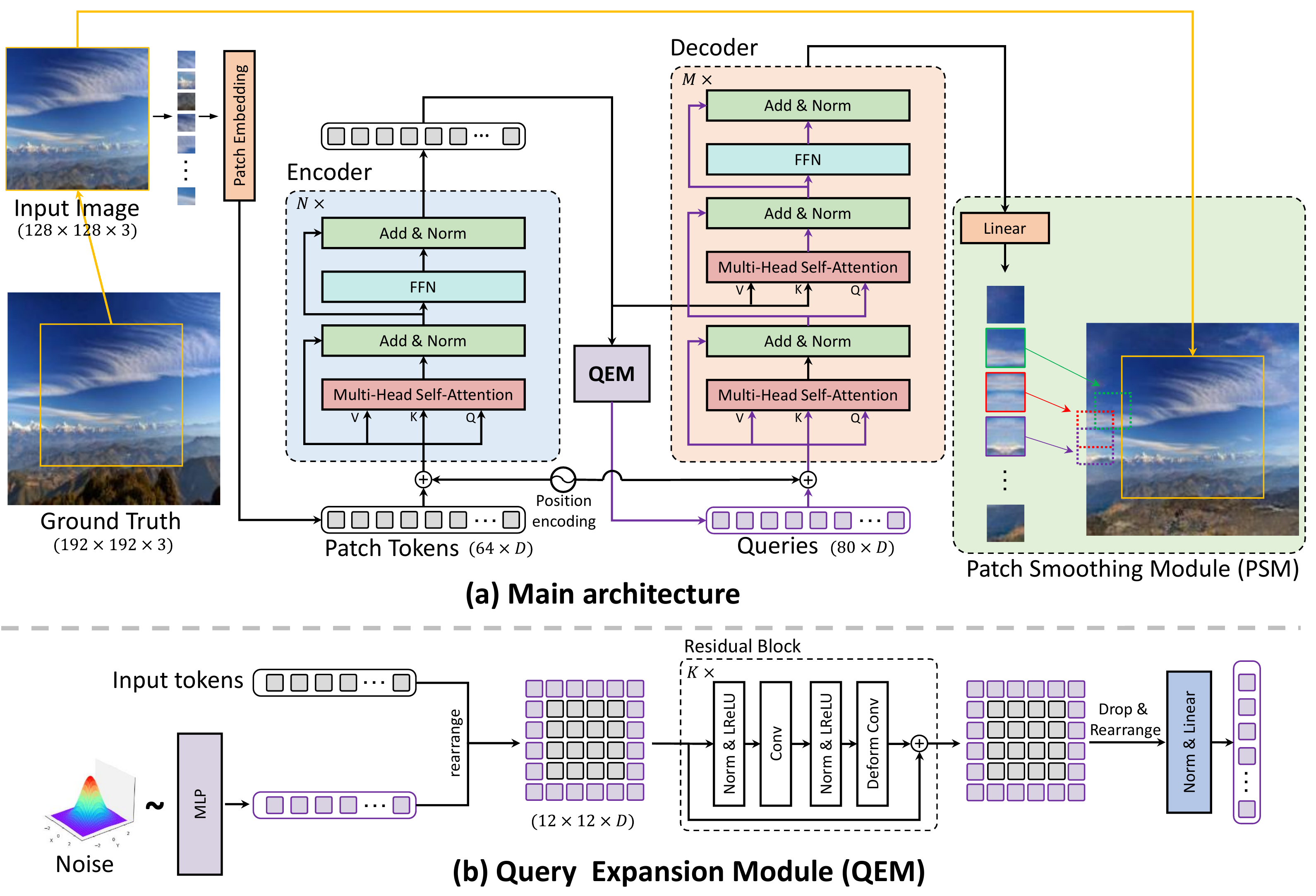}
\caption{(a) Main architecture of hybrid transformer generator in QueryOTR consists of transformer encoder and decoder, QEM and PSM. (b) Structure of Query Expansion Module.}
\label{fig:architecture}
\end{figure*}


\subsection{Problem Statement}
Given an image $\mathrm{x} \in \mathbb{R}^{H \times W \times 3}$, we aim to extrapolate  outside contents beyond the image boundary with extra $M$-pixels.
The generator will produce a visually convincing image $\hat{\mathrm{x}}\in \mathbb{R}^{(H+2M)\times (W+2M)\times 3}$. 
Different from previous work which is almost based on convolutional operations, we rephrase the problem as a patch-wise sequence-to-sequence autoregression problem.
In particular, we partition the image $\mathrm{x}$ into regular non-overlapping patches with the patch size $P \times P$ ($P$ is typically empirically set to 16), resulting in a sequence of patch tokens $\{\mathrm{x}_p^1,\mathrm{x}_p^2,\cdots,\mathrm{x}_p^L\}$, where $\mathrm{x}_p^i \in \mathbb{R}^{(P^2\cdot 3)}$ and the sequence length is $L=\frac{H \times W}{P^2}$. Our goal is to predict the extra sequence $\{\mathrm{x}_p^{L+1},\mathrm{x}_p^{L+2},\cdots,\mathrm{x}_p^{L+R}\}$ representing the extrapolated regions, where $\mathrm{x}_p^i \in \mathbb{R}^{(P^2\cdot 3)}$ and the expanded sequence length is $R=\frac{(H+2M)\times (W+2M)-H \times W}{P^2}$. The extrapolated image $\hat{\mathrm{x}}$ can be obtained by reshaping the new sequence of patch tokens into image patches, and then rearranging the image patches around the input image, leading to $\hat{\textbf{x}}=\mathcal{F}(\mathrm{x}, \{\mathrm{x}_p^{L+1},\mathrm{x}_p^{L+2},\cdots,\mathrm{x}_p^{L+R}\})$.

\subsection{Hybrid Transformer Autoencoder}
The architecture of the proposed QueryOTR generator is presented in \autoref{fig:architecture}, which is a hybrid transformer autoencoder. 
The overall architecture is composed of four major components: a transformer encoder extracting patch tokens' representation, a CNN-based Query Expansion Module (QEM) predicting the expanded queries, a transformer decoder processing the expanded queries, and a Patch Smoothing Module (PSM) generating the expanded patches and rearranging them around the original images. \\

\noindent\textbf{Transformer Encoder} Our encoder is a standard ViT~\cite{vit}. Inspired from ViT, the input image is first converted to several non-overlapping patches represented as a sequence of patch tokens $\mathrm{X}_p$. The encoder module embeds the patch tokens  through  a linear projection $\mathbf{E}$ with the added positional embeddings $\mathbf{E}_{pos}$. Then the encoder processes the set of patch tokens via a series of Transformer Blocks with a length of $N$. The transformer-based encoder can be described as follows:
\begin{align}
  &\mathrm{h}_0  = [\mathrm{x}_p^1\mathbf{E};\mathrm{x}_p^2\mathbf{E};...;\mathrm{x}_p^L\mathbf{E}] +\mathbf{E}_{pos}, && \mathbf{E}\in \mathbb{R}^{(P^2 \cdot 3)\times D},\mathbf{E}_{pos}\in \mathbb{R}^{L\times D}  \\
  &\mathrm{h}'_{n}  = \mathrm{MSA}(\mathrm{LN}(\mathrm{h}_{n-1}))+\mathrm{h}_{n-1}, && n=1,...,N \label{equ:2}\\
 & \mathrm{h}_{n}= \mathrm{FFN}(\mathrm{LN}(\mathrm{h}'_{n}))+\mathrm{h}'_{n}, && n=1,...,N\\
  &\mathrm{h}_{enc}= \mathrm{LN}(\mathrm{h}_{N}),&&
\end{align}
where $D$ is the hidden dimension of transformer block, $\mathrm{FFN}$ is a feed forward network, $\mathrm{LN}$ denotes layer normalization, $\mathrm{h}_n$ are the intermediate tokens' representations, $\mathrm{h}_{enc}$ denotes the output patch tokens of the transformer encoder, and MSA represents the multi-headed self-attention.

Given the learnable matrices $\mathbf{W}_q$, $\mathbf{W}_k$, $\mathbf{W}_v$ corresponding to query, key, and value representations, a single self-attention head (indexed with $h$) is computed:
\begin{equation}
\mathrm{Attention}_h(\mathbf{X},\mathbf{Y})=\mathrm{softmax}({\mathbf{Q}\mathbf{K}^\top}/{\sqrt{d_h}})\mathbf{V},
\end{equation}
where $\mathbf{Q}=\mathbf{X}\mathbf{W}_q$, $\mathbf{K}=\mathbf{Y}\mathbf{W}_k$, $\mathbf{V}=\mathbf{Y}\mathbf{W}_v$. Multi-headed self-attention aggregates information with linear projection operation on the concatenation of the $H$ self-attention heads:
\begin{align}
\mathrm{MSA}(\mathbf{X})=\mathrm{concat}_{h=1}^H [\mathrm{Attention}_h(\mathbf{X},\mathbf{X})]\mathbf{W}+\mathbf{b},
\end{align}
where $\mathbf{W}$ and $\mathbf{b}$ are learnable matrices for the aggregated features.\\

\noindent\textbf{Query Expansion Module} The proposed QEM is designed to speed up the convergence of pure transformer by generating the expanded queries for the transformer decoder. We predict the decoders' queries conditioned on encoders' features, and take advantage of CNN's inductive bias to accelerate the convergence.
As shown in \autoref{fig:architecture}(b), the input tokens $\mathrm{h}_{enc}$ are first reshaped to the feature map with the size of $\frac{H}{P} \times \frac{W}{P} \times D$. Then the reshaped feature maps are extrapolated with extra $\frac{M}{P}$ pixels along width and height, where the padded tokens are generated by Multi-layer Perceptual (MLP) with uniform input noise. After that, we utilize stacked residual blocks~\cite{resnet} equipped with deformable convolutional layers~\cite{dcn} to process the queries, which is commonly practiced to capture local and long-term dependencies. Finally, the expanded queries are extracted and transformed as sequence, followed by one Normalization Layer and one Linear Layer. This process can be described as:
\begin{align}
\mathrm{q}_{expand}&=\mathrm{QEM}(\mathrm{h}_{enc}, z),&&  z\sim \mathcal{N}(0,1).
\end{align}

\noindent\textbf{Transformer Decoder} Inspired from the original transformer~\cite{aiayn}, the decoder equips one extra sub-layer which performs the multi-head cross attention (MCA) similar to the encoder with two sub-layers. Specifically, in MCA the queries come from the previous decoder layer and the keys and values come from the output of the encoder. This allows each position in the decoder to attend over all positions in the input sequence, leading to significant improvements of the generating performance. The process can be described as follows:
\begin{align}
  \mathrm{q}_0 & = \mathrm{q}_{expand} +\mathbf{E}'_{pos}, &&  \mathbf{E}'_{pos}\in \mathbb{R}^{R\times D}\\
  \mathrm{q}'_{m} & = \mathrm{MSA}(\mathrm{LN}(\mathrm{q}_{m-1}))+\mathrm{q}_{m-1}, && m=1,...,M\\
  \mathrm{q}''_{m} & = \mathrm{MCA}(\mathrm{LN}(\mathrm{q}'_{m}), \mathrm{h}_{enc})+\mathrm{q}'_{m}, && m=1,...,M \label{eqn:mca}\\
  \mathrm{q}_{m}&= \mathrm{FFN}(\mathrm{LN}(\mathrm{q}''_{m}))+\mathrm{q}''_{m}, && m=1,...,M
\end{align}
The multi-headed cross-attention in \autoref{eqn:mca} aggregates information from $H$ cross attention heads, as follows:
\begin{align}
\mathrm{MCA}(\mathbf{X,Y})=\mathrm{concat}_{h=1}^H [\mathrm{Attention}_h(\mathbf{X},\mathbf{Y})]\mathbf{W}+\mathbf{b}.
\end{align}

\noindent\textbf{Patch Smoothing Module} The linear module is prone to generate artifacts if predicting output patches using predefined patch size of $P \times P$. The reason is that the output tokens are processed independently without explicit constraints. These arbitrary grid partitions could make the image contents discontinuous across the border edge of each patch. In order to mitigate this issue, we allow some overlaps among image patches. For each border edge of one patch, we extend it by $o$ pixels generating the output image patch size as $(P+2o) \times (P+2o)$. This operation involves the decoder with the neighboring patches' content having a better sense of locality in the transformer architecture, thus enabling the output sequence to have same length but less effect as the predefined grids. PSM can be described as:
\begin{align}
\hat{x} = &\mathcal{S}(x, \mathrm{q}_{M}\mathrm{W}_{proj}), && \mathrm{W}_{proj} \in \mathbb{R}^{D\times((P+2o)^2 \cdot 3)},
\end{align}
where $\mathcal{S}$ is a function to place the extrapolated overlapped patches around the input image, and average the pixel values in the overlapped areas.

\subsection{Loss Functions}
Our loss function consists of three parts: a patch-wise reconstruction loss, a perceptual loss, and an adversarial loss. The reconstruction loss is responsible for capturing the overall structure of predicted patches, whilst the perceptual loss and adversarial loss are coupled to maintain good perceptual quality and promote more realistic prediction.\\


\noindent\textbf{Patch-wise Reconstruction Loss} We utilize an L2 distance between the sequence of ground truth image patches $\{\mathrm{y}_p^{L+1},\mathrm{y}_p^{L+2},\cdots,\mathrm{y}_p^{L+R}\}$ and the sequence of predicted image patches $\{\mathrm{x}_p^{L+1},\mathrm{x}_p^{L+2},\cdots,\mathrm{x}_p^{L+R}\}$:
\begin{align}
\mathcal{L}_{rec} = \frac{1}{R}\sum\nolimits_{i=L+1}^{L+R} \Vert \mathrm{y}_p^{i}-\mathrm{x}_p^{i} \Vert_2^2,
\end{align}
where the patch size is $(P+2o) \times (P+2o)$. We engage a per-patch normalization to enhance the patch contrast locally, where the mean and std of the image patches are pre-computed.

\noindent\textbf{Perceptual Loss} Perceptual loss provides a supervision on the intermediate features that can help retain more semantic information. Following previous work~\cite{dosovitskiy2016generating,johnson2016perceptual,larsen2016autoencoding}, we extract the features from a VGG-19~\cite{vgg19} network pretrained on ImageNet~\cite{deng2009imagenet}, which is denoted as $\phi$. The perceptual loss is devised as follows:
\begin{equation}
\mathcal{L}_{perceptual} = \frac{1}{5}\sum\nolimits_{j=1}^{5}(w^j\times (\phi^j(\hat{x})-\phi^j(y))),
\end{equation}
where the superscript $j$ is the index of feature map scales from $\phi$, and $w^j$ is set to $1/32, 1/16, 1/8, 1/4, 1$ as the scale decreases.\\

\noindent\textbf{Adversarial Loss} We utilize the same multi-scale PatchGAN discriminator $D$ used in pix2pixHD~\cite{pix2pixhd} except that we replace the least squared loss term~\cite{lsgan} with the hinge loss term~\cite{hingegan}. Since the PatchGAN discriminator has a fixed receptive field of patch, we take the whole generated images instead of image patches to train the GAN. The extrapolated images generated by our QueryOTR should be indistinguishable from real images by the discriminator. Given the extrapolated images $\mathrm{\hat{x}}\sim\mathbb{P}_g$ generated by  QueryOTR   and real images $\mathrm{y}\sim\mathbb{P}_y$, the adversarial loss for the discriminator is
\begin{align}
\mathcal{L}_{adv}^{D} = \min\limits_{D} \mathbb{E}_{\mathrm{\hat{x}}\sim\mathbb{P}_g}(min(1+D(\hat{x})))+\mathbb{E}_{\mathrm{y}\sim\mathbb{P}_y}(min(1-D(y))).
\end{align}
Additionally, the adversarial loss for the generator is
\begin{align}
\mathcal{L}_{adv}^{G} = \min\limits_{G}  -\mathbb{E}_{\mathrm{\hat{x}}\sim\mathbb{P}_g}D(\hat{x}).
\end{align}
We jointly train the hybrid transformer generator and CNN discriminators and optimize the final objective as a weighted sum of the above mentioned loss terms:
\begin{align}
\min\limits_{G} \max\limits_{D} \mathcal{L}_{adv}+\lambda_{rec}\mathcal{L}_{rec}+\lambda_{perceptual}\mathcal{L}_{perceptual},
\end{align}
where $\lambda_{rec}$,  $\lambda_{perceptual}$ are weights controlling the importance of loss terms. In our experiments, we set $\lambda_{rec}=5$, and $\lambda_{perceptual}=10$.


\section{Experiments}



\subsection{Datasets, Implementation and Training Details}
We use three datasets with \{Scenery~\cite{NSIPO}, Building Facades~\cite{UTransformer}, and WikiArt~\cite{wikiart}\} for the experiments. Details about the three datasets could be found in the supplementary materials.

We implement our framework with PyTorch~\cite{pytorch} equipped with a NVIDIA GeForce RTX 3090 GPU 1.9.0. Hybrid transformer generator contains 12 stacked transformer encoder layers and 4 stacked transformer decoder layers. We initialise the weights of generator encoder by utilizing the pre-trained ViT~\cite{mae}. Adam~\cite{adam} is used as the optimizer to minimize the objective function with the mini-batch of 64, $\beta_1=0.0$, $\beta_2=0.99$, and weight decay of 0.0001. The $o$ is set to 8 considering the complexity and precision. Our QueryOTR is trained for 300, 200 and 120 epochs on Scenery, Building Facades, and WikiArt datasets respectively with the learning rate of 0.0001.
The warm-up trick~\cite{resnet} is utilized in the first 10 epochs with the reconstruction loss only.
For discriminator regularization, DiffAug~\cite{diffaug} and spectral normalization~\cite{sngan} are used to stabilise the training dynamics.

We conduct generalised image outpainting for experimental comparison following the previous work.
In the training stage, the original images are resized to the size $192\times192$ as the ground truth images. Then the input images with the size $128\times128$ are obtained by the center cropping operation. In the testing stage, all images are resized to $192\times192$ as the ground truth, and then the input images are obtained by center cropping to the sizes $128\times128$, $86\times86$, and $56\times56$ for $1\times$, $2\times$, and $3\times$ outpainting respectively. Excepted for horizontal flip and image normalization, no other data augmentation is used for ease of setup. The total output sizes are 2.25, 5, and 11.7 times of the input in terms of $1\times$, $2\times$, and $3\times$ outpainting, indicating that over half of all pixels will be generated.



\subsection{Experimental Results}

\begin{table}[!t]
\setlength\tabcolsep{1.5pt}
\centering
\begin{tabular}{c|l|ccc|ccc|ccc|}
\cline{2-11}
                                                 & \multirow{2}{*}{Methods} & \multicolumn{3}{c|}{Scenery}                                                                          & \multicolumn{3}{c|}{Building Facades}                                                                 & \multicolumn{3}{c|}{WikiArt}                                                              \\ \cline{3-11}
                                                 &                          & \multicolumn{1}{c|}{FID$\downarrow$}    & \multicolumn{1}{c|}{IS$\uparrow$}      & PSNR$\uparrow$     & \multicolumn{1}{c|}{FID$\downarrow$}    & \multicolumn{1}{c|}{IS$\uparrow$}      & PSNR$\uparrow$     & \multicolumn{1}{c|}{FID$\downarrow$} & \multicolumn{1}{c|}{IS$\uparrow$} & PSNR$\uparrow$ \\ \hline
\multicolumn{1}{|c|}{\multirow{5}{*}{$1\times$}} & SRN                      & \multicolumn{1}{c|}{47.781}             & \multicolumn{1}{c|}{2.981}             & 22.440             & \multicolumn{1}{c|}{38.644}             & \multicolumn{1}{c|}{3.862}             & 18.588             & \multicolumn{1}{c|}{76.749}                & \multicolumn{1}{c|}{3.629}             &      \underline{20.072}          \\
\multicolumn{1}{|c|}{}                           & NSIPO                    & \multicolumn{1}{c|}{25.977}             & \multicolumn{1}{c|}{3.059}             & 21.089             & \multicolumn{1}{c|}{30.465}             & \multicolumn{1}{c|}{4.153}             & 18.314             & \multicolumn{1}{c|}{22.242}                & \multicolumn{1}{c|}{5.600}             &    18.592            \\
\multicolumn{1}{|c|}{}                           & IOH                      & \multicolumn{1}{c|}{32.107}             & \multicolumn{1}{c|}{2.886}             & 22.286             & \multicolumn{1}{c|}{49.481}             & \multicolumn{1}{c|}{3.924}             & 18.431             & \multicolumn{1}{c|}{40.184}                & \multicolumn{1}{c|}{4.835}             &     19.403           \\
\multicolumn{1}{|c|}{}                           & Uformer                  & \multicolumn{1}{c|}{\underline{20.575}} & \multicolumn{1}{c|}{\underline{3.249}} & \underline{23.007} & \multicolumn{1}{c|}{\underline{30.542}} & \multicolumn{1}{c|}{\underline{4.189}} & \underline{18.828} & \multicolumn{1}{c|}{\underline{15.904}}          & \multicolumn{1}{c|}{\underline{6.567}}        &    19.610      \\
\multicolumn{1}{|c|}{}                           & QueryOTR                 & \multicolumn{1}{c|}{\textbf{20.366}}    & \multicolumn{1}{c|}{\textbf{3.955}}    & \textbf{23.604}    & \multicolumn{1}{c|}{\textbf{22.378}}    & \multicolumn{1}{c|}{\textbf{4.978}}    & \textbf{19.680}    & \multicolumn{1}{c|}{\textbf{14.955}}                & \multicolumn{1}{c|}{\textbf{7.896}}             & \textbf{20.388}               \\ \hline
\multicolumn{1}{|c|}{\multirow{5}{*}{2$\times$}} & SRN                      & \multicolumn{1}{c|}{83.772}             & \multicolumn{1}{c|}{2.349}             & 18.403             & \multicolumn{1}{c|}{74.304}             & \multicolumn{1}{c|}{3.651}             & 15.355             & \multicolumn{1}{c|}{137.997}                & \multicolumn{1}{c|}{3.039 }              &      \underline{16.646}        \\
\multicolumn{1}{|c|}{}                           & NSIPO                    & \multicolumn{1}{c|}{45.989}             & \multicolumn{1}{c|}{2.606}             & 17.733             & \multicolumn{1}{c|}{\underline{58.341}} & \multicolumn{1}{c|}{3.669}             & 15.262             & \multicolumn{1}{c|}{51.668}                & \multicolumn{1}{c|}{4.591}             &     15.679           \\
\multicolumn{1}{|c|}{}                           & IOH                      & \multicolumn{1}{c|}{44.742}             & \multicolumn{1}{c|}{2.655}             & 18.739             & \multicolumn{1}{c|}{76.476}             & \multicolumn{1}{c|}{3.456}             & 15.443             & \multicolumn{1}{c|}{75.070}                & \multicolumn{1}{c|}{4.289}             &      16.056          \\
\multicolumn{1}{|c|}{}                           & Uformer                  & \multicolumn{1}{c|}{\underline{39.801}} & \multicolumn{1}{c|}{\underline{2.920}} & \underline{18.920} & \multicolumn{1}{c|}{63.915}             & \multicolumn{1}{c|}{\underline{3.798}} & \underline{15.612} & \multicolumn{1}{c|}{\textbf{41.107}}                & \multicolumn{1}{c|}{\underline{5.900}}             &        15.947        \\
\multicolumn{1}{|c|}{}                           & QueryOTR                 & \multicolumn{1}{c|}{\textbf{39.237}}    & \multicolumn{1}{c|}{\textbf{3.431}}    & \textbf{19.358}    & \multicolumn{1}{c|}{\textbf{41.273}}    & \multicolumn{1}{c|}{\textbf{4.547}}    & \textbf{16.213}    & \multicolumn{1}{c|}{\underline{43.757}}                & \multicolumn{1}{c|}{\textbf{6.341}}             &   \textbf{17.074}           \\ \hline
\multicolumn{1}{|c|}{\multirow{5}{*}{3$\times$}} & SRN                      & \multicolumn{1}{c|}{115.193}            & \multicolumn{1}{c|}{2.087}             & 16.123             & \multicolumn{1}{c|}{110.036}            & \multicolumn{1}{c|}{2.938}             & 13.693             & \multicolumn{1}{c|}{181.533}                & \multicolumn{1}{c|}{2.504}             &      \underline{14.609}          \\
\multicolumn{1}{|c|}{}                           & NSIPO                    & \multicolumn{1}{c|}{64.457}             & \multicolumn{1}{c|}{2.405}             & 15.606             & \multicolumn{1}{c|}{\underline{81.301}} & \multicolumn{1}{c|}{\underline{3.431}} & 13.791             & \multicolumn{1}{c|}{75.785}                & \multicolumn{1}{c|}{4.225}             &        14.257        \\
\multicolumn{1}{|c|}{}                           & IOH                      & \multicolumn{1}{c|}{\textbf{58.629}}    & \multicolumn{1}{c|}{2.432}             & 16.307             & \multicolumn{1}{c|}{95.068}             & \multicolumn{1}{c|}{2.790}             & 13.894             & \multicolumn{1}{c|}{108.328}                & \multicolumn{1}{c|}{3.728}             &      13.919          \\
\multicolumn{1}{|c|}{}                           & Uformer                  & \multicolumn{1}{c|}{\underline{60.497}} & \multicolumn{1}{c|}{\underline{2.638}} & \underline{16.379} & \multicolumn{1}{c|}{93.888}             & \multicolumn{1}{c|}{3.388}             & \underline{14.051} & \multicolumn{1}{c|}{\underline{72.923}}                & \multicolumn{1}{c|}{\textbf{5.904}}             &      13.464          \\
\multicolumn{1}{|c|}{}                           & QueryOTR                 & \multicolumn{1}{c|}{60.977}             & \multicolumn{1}{c|}{\textbf{3.114}}    & \textbf{16.864}    & \multicolumn{1}{c|}{\textbf{64.926}}    & \multicolumn{1}{c|}{\textbf{4.612}}    & \textbf{14.316}    & \multicolumn{1}{c|}{\textbf{69.951}}                & \multicolumn{1}{c|}{\underline{5.683}}             &   \textbf{15.294}             \\ \hline
\end{tabular}
\caption{Quantitative results of one-step and multi-step outpainting. Best and second best results are \textbf{boldface} and \underline{underlined}. $1 \times$ represents one step outpainting, while $2 \times$ and $3 \times$ denote two- and three-step outpainting respectively.}
\label{tab:comparison}
\end{table}

We make comparisons with three SOTA CNN-based image outpainting methods, NSIPO~\cite{NSIPO}, SRN~\cite{SRN}, and IOH~\cite{IOHGan}, and one transformer-based method Uformer~\cite{UTransformer} to demonstrate the effectiveness of QueryOTR. For all the experiments, we set the input and output sizes as $128\times128$ and $192\times192$.

We use Inception Score (IS)~\cite{IS}, Fr\'echet Inception Distance (FID)~\cite{FIDTTUR}, and peak signal-to-noise ratio (PSNR) to measure the generative quality objectively.
The upper-bounds of IS are 4.091, 5.660 and 8.779 for Scenery, Building Facades and WikiArt, respectively, which are calculated by real images in test set.

\begin{figure}[!t]
\centering
\includegraphics[width=0.96\columnwidth]{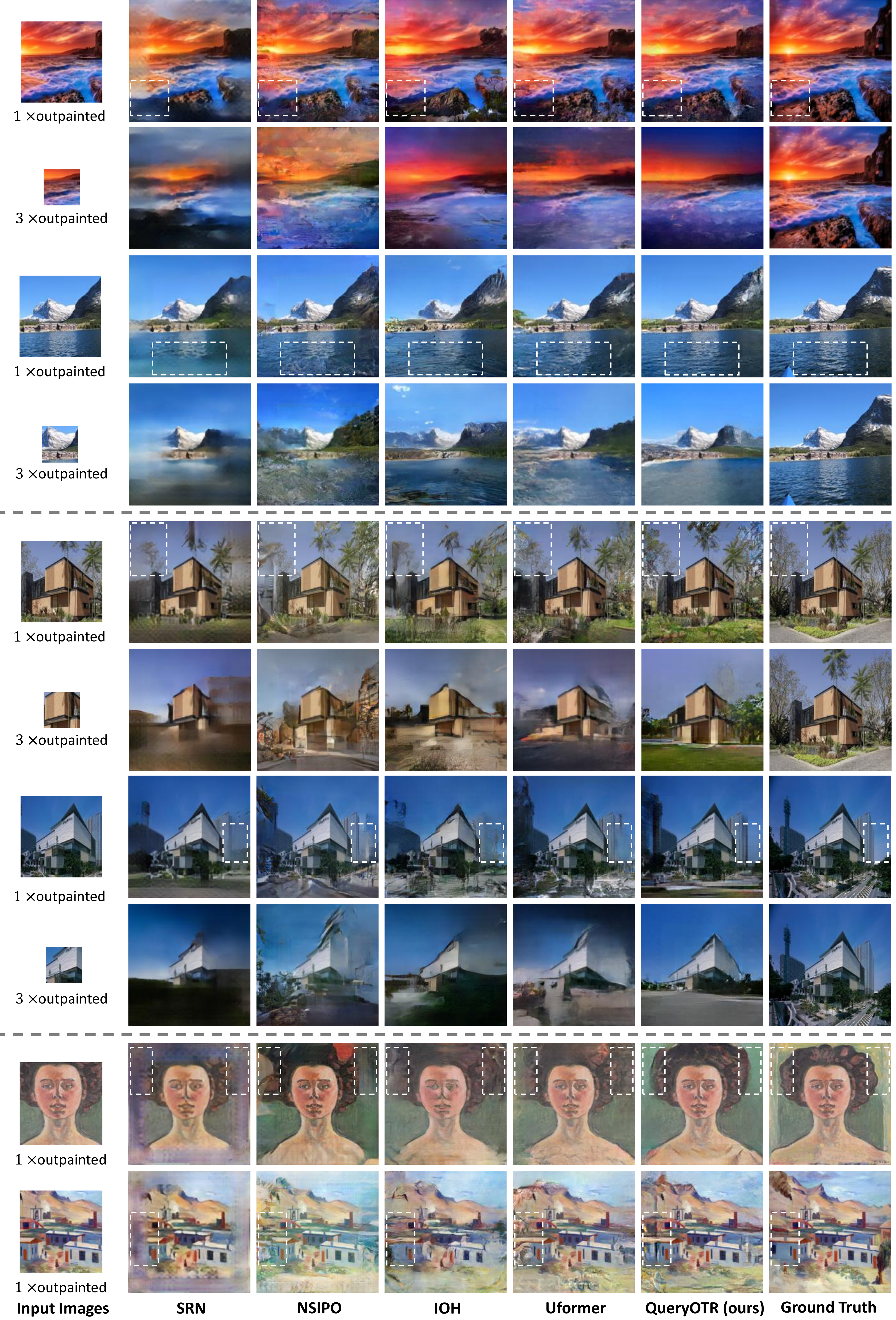}
\caption{Comparisons on 1-step and multi-step outpainting with the state-of-the-art methods. Our QueryOTR achieves the best image quality.}
\label{fig:compare}
\end{figure}

\noindent\textbf{Quantitative Result} \autoref{tab:comparison} shows quantitative results. Our QueryOTR outperforms the competition on almost all metrics on 1-step and multi-step outpainting. In particular,  QueryOTR shows obvious superiority in all entries compared with CNN-based methods, e.g., SRN, NSIPO, and IOH. These results show that transformer structure succeeds in capturing global dependencies for image outpainting compared with CNN's inductive biases. Meanwhile, our  QueryOTR outperforms the very competitive Swin-based Uformer which uses an image-to-image translation approach for image extrapolation, mainly because our query-based method allows to generate image patches attended to all the visual locations, yielding a better perceptual consistency. It is noted that our results for $1\times$ outpainting are very close to the IS upper-bound for all the datasets, indicating realistic image generation and good perceptual consistency. Extra results of replacing the center region with input sub-images are in the supplementary.

\noindent\textbf{Qualitative Result} Some examples of visual results on all the datasets are shown in \autoref{fig:compare}. Our QueryOTR effectively extrapolates the images by  querying the global semantic-similar image patches. Seen from the $1\times$ outpainting results, our QueryOTR could generate more realistic images with vivid details and enrich the contents of the generated regions marked in white box. Furthermore, our method could weaken the sense of edges between the generated regions and input sub-image. Compared with other baselines, our QueryOTR could generate water containing more realistic ripples in the $3^{rd}$ row and intact trees in the $5^{th}$ row of \autoref{fig:compare}, which could be seen in the white dotted box. In the $7^{th}$ row of \autoref{fig:compare}, the whole skyscraper generated by QueryOTR indicates the success of our query-based method which predicts the detailed contents with global information by queries. In the $9^{th}$ row, our method could capture the global information of the green background on the corner marked in the white box. More visual results could be seen in the supplementary material.

\subsection{Ablation Study}
We ablate several critical factors in QueryOTR by progressively adjusting each factor here. It can be seen that each factor contributes to the final success of QueryOTR. We conducted all the ablation experiments on the Scenery dataset.\\

\noindent\textbf{Transformer Encoder and Decoder} We compare the impact of the pretrained ViT-based encoder and the number of transformer decoder layers $M$. As shown in \autoref{tab:ablation}(a), utilizing a pretrained ViT encoder contributes to the improvements of FID and IS by $2.418$ and $0.204$, respectively. The main reason is that the small datasets might not be sufficient to train the model for performance saturation. The pretrained ViT encoder is capable of capturing the long-term dependencies, which may benefit the patch prediction. Additionally, our QueryOTR performs optimally in both FID and IS when the number of decoder layers is set to 4. Further increasing the depth of decoder indefinitely will not improve the performance of our QueryOTR.

\begin{table}[!t]
\setlength\tabcolsep{2.5pt}
  \begin{minipage}[t]{.48\linewidth}
    \centering
    \subfloat[][Ablation of the pretrained ViT-base encoder and the number of transformer decoder layers $M$.]{\label{ablation:a}
    \begin{tabular}{c|c|cc}
      \hline
  Pretrained \textit{Enc.} & $M$ & FID$\downarrow$ & IS$\uparrow$ \\\hline
  -  &   4 &    22.784      &  3.751   \\
  \checkmark  &  2 &  20.731  & 3.931\\
  \checkmark &    4 &    \textbf{20.366}  &  \textbf{3.955}     \\
  \checkmark &    8 &   20.373  &  3.852 \\
      \hline
    \end{tabular}}
  \end{minipage}\hfill
  \begin{minipage}[t]{.48\linewidth}
    \centering
    \subfloat[][Impact of $\mathcal{L}_{rec}$ and $\mathcal{L}_{perceptual}$ contribute to the overall performance. The model is default trained with three losses.]{\label{ablation:b}
    \begin{tabular}{l|cc}
      \hline
     & FID$\downarrow$ & IS$\uparrow$ \\\hline
      w/o $\mathcal{L}_{rec}$ \& $\mathcal{L}_{perceptual}$  & 38.009 &  3.433 \\
       w/o $\mathcal{L}_{rec}$ &  31.282   & 3.744   \\
       w/o $\mathcal{L}_{perceptual}$ &   33.380 &  3.510\\
       QueryOTR (baseline) &   \textbf{20.366}  &  \textbf{3.955} \\
      \hline
    \end{tabular}}
  \end{minipage}\hfill
  \quad
    \begin{minipage}[t]{.48\linewidth}
    \centering
    \subfloat[][Impact of proposed Query Expansion Module (QEM) and its key internal components.]{\label{ablation:c}     
    \begin{tabular}{l|cc}
      \hline
     & FID$\downarrow$ & IS$\uparrow$ \\\hline
       w/o QEM &  36.967   &3.642    \\
       QEM w/o  Noise&  23.444  &  3.728 \\
       QEM w/o DC~\cite{dcn} &23.530 & \textbf{3.775}   \\
        w QEM &   \textbf{22.784}  & 3.751 \\
      \hline
    \end{tabular}}
  \end{minipage}\hfill
  \begin{minipage}[t]{.48\linewidth}
    \centering
    \subfloat[][Effect of the proposed Patch Smoothing Module (PSM) and per-patch image normalization.]{\label{ablation:d}
    \begin{tabular}{c|c|cc}
      \hline
     PSM & Per-Patch \textit{Norm.}  & FID$\downarrow$ & IS$\uparrow$ \\\hline
      - &- &  51.945   &  3.801  \\
     - &\checkmark &  31.073  &  3.753  \\
      \checkmark & - &  22.501   & 3.707   \\
      \checkmark &\checkmark &   \textbf{20.366}  &  \textbf{3.955} \\
      \hline
    \end{tabular}}
  \end{minipage}\hfill
\caption{Ablation studies validated on Scenery dataset.}
\label{tab:ablation}
\end{table}

\begin{figure}[t]
\centering
\includegraphics[width=\columnwidth]{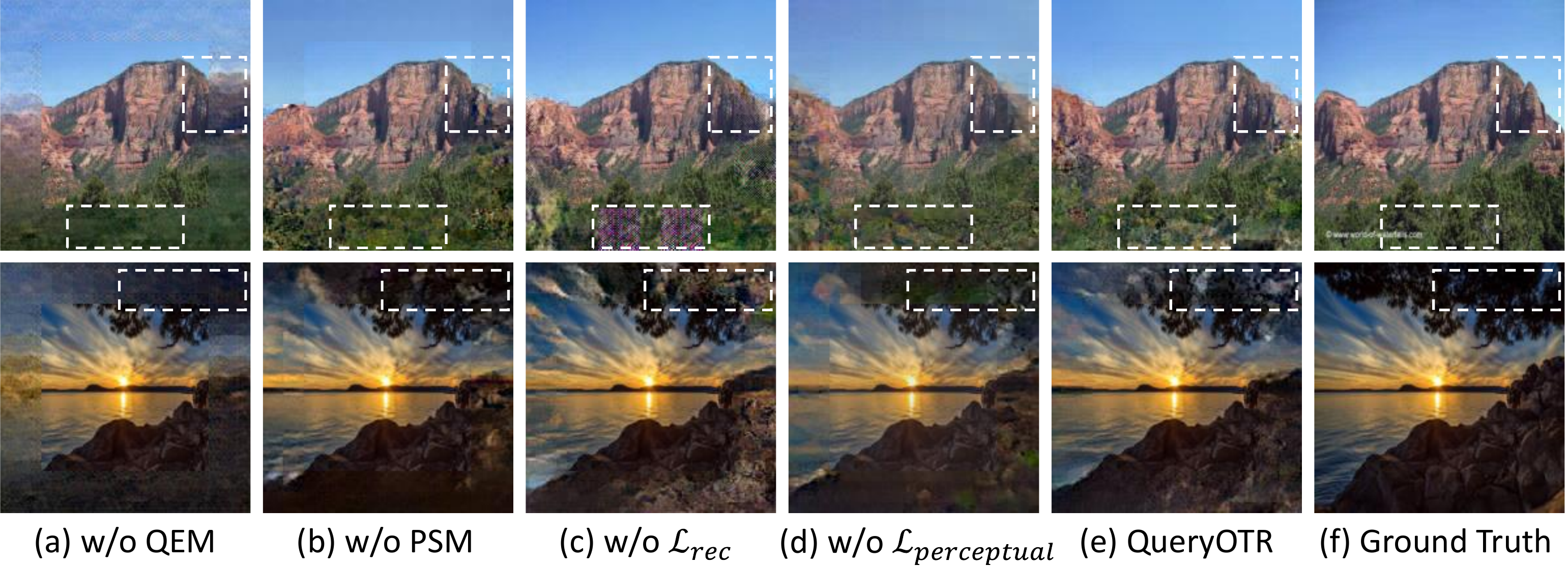}
\caption{Visualisation of ablation study.}
\label{fig:ablation}
\end{figure}

\noindent\textbf{Loss Terms} We investigate the impact of patch-wise reconstruction loss $\mathcal{L}_{rec}$ and perceptual loss $\mathcal{L}_{perceptual}$ in \autoref{tab:ablation}(b). We first train the model with only adversarial loss, which is equivalent to training the model unpaired, resulting in a FID of $38.009$ and IS of $3.433$. On the basis of adversarial training, using either $\mathcal{L}_{rec}$ or $\mathcal{L}_{perceptual}$ could improve the overall performance.
\autoref{fig:ablation}(c) and (d) show that high-frequency checkerboard artifacts occur when trained without $\mathcal{L}_{rec}$, and the details cannot be generated without $\mathcal{L}_{perceptual}$. 

\noindent\textbf{QEM} We ablate the impact of QEM and its internal key components. In the experiments, we do not use a pretrained encoder to avoid reducing the difficulty of training learnable queries. Since training pure transformer may require larger datasets and longer time, it is hard for learnable queries to converge well on Scenery dataset, resulting in a high FID (see \autoref{tab:ablation}(c)) and blurry image patches (see \autoref{fig:ablation}(a)). On the other hand, the proposed QEM generates queries conditioned on input images, significantly improving  FID by 14.227. Meanwhile, generating queries with noise slightly improves the patch diversity, and deformable convolution enables an active long distance modeling for query generation.

To further investigate how QEM affect the convergence speed of pure transformer, we train the pure transformer with and without QEM module for 1000 epochs. As shown in \autoref{vis}(a), the convergence rate of the pure transformer with QEM is about 3.3 times faster than that without QEM on a relatively small dataset indicating the superiority of QEM in accelerating the model convergence. On the other hand, the loss declines slowly without QEM, which might be caused by the insufficient training data. The reason leading to this phenomenon is that the pure transformer will process almost 4 billion possibilities if the $16\times 16$ pixel patch is treated as a word, which needs larger semantic space for attention processing. When dealing with a small dataset, the amount of data is not enough to regress the extrapolated patches resulting in model degradation.


\noindent\textbf{PSM} \autoref{tab:ablation}(d) demonstrates the effect of the proposed PSM and per-patch normalization. Although using a single linear layer can generate vivid image patches, the connections between patches are unnatural, as shown in \autoref{fig:ablation}(b). Per-patch normalization could improve the reconstruction of high-frequency by enhancing the local contrast of patches, leading an improvement of FID 20.872. Meanwhile, PSM significantly alleviates the checkerboard artifacts caused by per-patch prediction, and improves the overall perceptual quality of the extrapolated images.
PSM alleviates checkerboard artifacts via explicit constraints, while perceptual loss penalizes image discontinuity from a semantic perspective. PSM appears more effective and direct than perceptual loss. If both are applied, even better performance can be obtained.

\section{Conclusion}
In this paper, we have proposed a novel hybrid query-based encoder-decoder transformer framework, \textbf{QueryOTR}, to extrapolate visual context all-side around a given image. The transformer structure breaks through the limitation of capturing image long-rang dependencies and intrinsic locality. 
The special designed module QEM helps to accelerate the transformer model convergence on small datasets and PSM contributes to generate seamless extrapolated images realistically and smoothly. Extensive experiments on Scenery, Building and WikiArt datasets proved the superiority of our query-based method.\\

\noindent\textbf{Acknowledgments.}
The work was partially supported by the following: National Natural Science Foundation of China under no.61876155; Jiangsu Science and Technology Programme under no.BE2020006-4; Key Program Special Fund in XJTLU under no.KSF-T-06, no.KSF-E-26 and no.KSF-E-37; Research Development Fund in XJTLU under no.RDF-19-01-21.

\clearpage
%
%
\bibliographystyle{splncs04}
\bibliography{egbib}

\clearpage

\appendix

\section{Additional Quantitative Results}

\begin{table}[!h]
\setlength\tabcolsep{1.5pt}
\centering
\begin{tabular}{c|l|ccc|ccc|ccc|}
\cline{2-11}
\multicolumn{1}{l|}{\multirow{2}{*}{}} & \multicolumn{1}{c|}{\multirow{2}{*}{Methods}} & \multicolumn{3}{c|}{Scenery} & \multicolumn{3}{c|}{Building Facades} & \multicolumn{3}{c|}{WikiArt} \\ \cline{3-11}
\multicolumn{1}{l|}{} & \multicolumn{1}{c|}{} & \multicolumn{1}{c|}{FID$\downarrow$} & \multicolumn{1}{c|}{IS$\uparrow$} & PSNR$\uparrow$ & \multicolumn{1}{c|}{FID$\downarrow$} & \multicolumn{1}{c|}{IS$\uparrow$} & PSNR$\uparrow$ & \multicolumn{1}{c|}{FID$\downarrow$} & \multicolumn{1}{c|}{IS$\uparrow$} & PSNR$\uparrow$ \\ \hline
\multicolumn{1}{|c|}{\multirow{6}{*}{$1\times$}} & Lower Bound & \multicolumn{1}{c|}{160.174} & \multicolumn{1}{c|}{3.595} &9.569 & \multicolumn{1}{c|}{123.678} & \multicolumn{1}{c|}{4.356} & 9.810 & \multicolumn{1}{c|}{139.956} & \multicolumn{1}{c|}{5.073} &10.215 \\
\multicolumn{1}{|c|}{} & SRN & \multicolumn{1}{c|}{45.296} & \multicolumn{1}{c|}{3.540} & 22.433 & \multicolumn{1}{c|}{34.058} & \multicolumn{1}{c|}{4.722} & 18.839 & \multicolumn{1}{c|}{65.675} & \multicolumn{1}{c|}{4.933} & \textbf{20.467} \\
\multicolumn{1}{|c|}{} & NSIPO & \multicolumn{1}{c|}{35.606} & \multicolumn{1}{c|}{3.475} & 21.630 & \multicolumn{1}{c|}{33.140} & \multicolumn{1}{c|}{4.529} & 18.460 & \multicolumn{1}{c|}{30.338} & \multicolumn{1}{c|}{6.231} & 18.929 \\
\multicolumn{1}{|c|}{} & IOH & \multicolumn{1}{c|}{23.410} & \multicolumn{1}{c|}{3.578} & 22.839 & \multicolumn{1}{c|}{33.525} & \multicolumn{1}{c|}{\underline{4.739}} & 18.812 & \multicolumn{1}{c|}{24.539} & \multicolumn{1}{c|}{6.679} & 19.808 \\
\multicolumn{1}{|c|}{} & Uformer & \multicolumn{1}{c|}{\underline{23.216}} & \multicolumn{1}{c|}{\underline{3.691}} & \underline{23.054} & \multicolumn{1}{c|}{\underline{32.228}} & \multicolumn{1}{c|}{4.651} & \underline{18.892} & \multicolumn{1}{c|}{\underline{18.808}} & \multicolumn{1}{c|}{\underline{7.466}} & 19.708 \\
\multicolumn{1}{|c|}{} & QueryOTR & \multicolumn{1}{c|}{\textbf{20.366}} & \multicolumn{1}{c|}{\textbf{3.955}} & \textbf{23.604} & \multicolumn{1}{c|}{\textbf{22.378}} & \multicolumn{1}{c|}{\textbf{4.978}} & \textbf{19.680} & \multicolumn{1}{c|}{\textbf{14.955}} & \multicolumn{1}{c|}{\textbf{7.896}} & \underline{20.388} \\ \hline
\multicolumn{1}{|c|}{\multirow{6}{*}{$2\times$}} & Lower Bound & \multicolumn{1}{c|}{201.871} & \multicolumn{1}{c|}{2.097} &7.868 & \multicolumn{1}{c|}{196.650} & \multicolumn{1}{c|}{2.875} & 8.047 & \multicolumn{1}{c|}{230.893} & \multicolumn{1}{c|}{2.477} &8.557 \\
\multicolumn{1}{|c|}{} & SRN & \multicolumn{1}{c|}{97.989} & \multicolumn{1}{c|}{2.724} & 18.459 & \multicolumn{1}{c|}{75.121} & \multicolumn{1}{c|}{3.837} & 15.431 & \multicolumn{1}{c|}{139.395} & \multicolumn{1}{c|}{3.045} & \underline{16.759} \\
\multicolumn{1}{|c|}{} & NSIPO & \multicolumn{1}{c|}{69.683} & \multicolumn{1}{c|}{\underline{3.235}} & 17.701 & \multicolumn{1}{c|}{\underline{ 65.319}} & \multicolumn{1}{c|}{3.771} & 15.287 & \multicolumn{1}{c|}{67.880} & \multicolumn{1}{c|}{4.888} & 15.721 \\
\multicolumn{1}{|c|}{} & IOH & \multicolumn{1}{c|}{\underline{45.108}} & \multicolumn{1}{c|}{3.047} & 18.846 & \multicolumn{1}{c|}{72.053} & \multicolumn{1}{c|}{3.727} & 15.519 & \multicolumn{1}{c|}{66.953} & \multicolumn{1}{c|}{5.065} & 16.127 \\
\multicolumn{1}{|c|}{} & Uformer & \multicolumn{1}{c|}{50.605} & \multicolumn{1}{c|}{3.099} & \underline{18.934} & \multicolumn{1}{c|}{71.306} & \multicolumn{1}{c|}{\underline{3.924}} & \underline{15.626} & \multicolumn{1}{c|}{\underline{51.263}} & \multicolumn{1}{c|}{\underline{6.098}} & 15.936 \\
\multicolumn{1}{|c|}{} & QueryOTR & \multicolumn{1}{c|}{\textbf{39.237}} & \multicolumn{1}{c|}{\textbf{3.431}} & \textbf{19.358} & \multicolumn{1}{c|}{\textbf{41.273}} & \multicolumn{1}{c|}{\textbf{4.547}} & \textbf{16.213} & \multicolumn{1}{c|}{\textbf{43.757}} & \multicolumn{1}{c|}{\textbf{6.341}} & \textbf{17.074} \\ \hline
\multicolumn{1}{|c|}{\multirow{6}{*}{$3\times$}} & Lower Bound & \multicolumn{1}{c|}{227.268} & \multicolumn{1}{c|}{1.991} &7.242 & \multicolumn{1}{c|}{223.224} & \multicolumn{1}{c|}{2.378} & 7.384 & \multicolumn{1}{c|}{260.623} & \multicolumn{1}{c|}{2.258} & 7.919\\
\multicolumn{1}{|c|}{} & SRN & \multicolumn{1}{c|}{141.040} & \multicolumn{1}{c|}{2.483} & 16.141 & \multicolumn{1}{c|}{114.016} & \multicolumn{1}{c|}{3.312} & 13.777 & \multicolumn{1}{c|}{181.394} & \multicolumn{1}{c|}{2.407} & \underline{14.620} \\
\multicolumn{1}{|c|}{} & NSIPO & \multicolumn{1}{c|}{101.411} & \multicolumn{1}{c|}{\textbf{3.131}} & 15.384 & \multicolumn{1}{c|}{\underline{92.041}} & \multicolumn{1}{c|}{\underline{3.628}} & 13.741 & \multicolumn{1}{c|}{94.176} & \multicolumn{1}{c|}{4.325} & 14.159 \\
\multicolumn{1}{|c|}{} & IOH & \multicolumn{1}{c|}{\underline{67.591}} & \multicolumn{1}{c|}{2.723} & 16.351 & \multicolumn{1}{c|}{104.337} & \multicolumn{1}{c|}{2.956} & 13.913 & \multicolumn{1}{c|}{104.032} & \multicolumn{1}{c|}{4.190} & 13.943 \\
\multicolumn{1}{|c|}{} & Uformer & \multicolumn{1}{c|}{76.318} & \multicolumn{1}{c|}{2.799} & \underline{16.374} & \multicolumn{1}{c|}{105.539} & \multicolumn{1}{c|}{3.315} & \underline{14.065} & \multicolumn{1}{c|}{\underline{79.322}} & \multicolumn{1}{c|}{\textbf{5.954}} & 13.411 \\
\multicolumn{1}{|c|}{} & QueryOTR & \multicolumn{1}{c|}{\textbf{60.977}} & \multicolumn{1}{c|}{\underline{3.114}} & \textbf{16.864} & \multicolumn{1}{c|}{\textbf{64.926}} & \multicolumn{1}{c|}{\textbf{4.612}} & \textbf{14.316} & \multicolumn{1}{c|}{\textbf{69.951}} & \multicolumn{1}{c|}{5.683} & \textbf{15.294} \\ \hline
\multicolumn{1}{|c|}{\multirow{1}{*}{}} & Up Bound & \multicolumn{1}{c|}{0} & \multicolumn{1}{c|}{4.184} & +$\infty$ & \multicolumn{1}{c|}{0} & \multicolumn{1}{c|}{5.660} &+$\infty$  & \multicolumn{1}{c|}{0} & \multicolumn{1}{c|}{8.779} & +$\infty$\\\hline
\end{tabular}
\caption{Quantitative results follow our setting that \textbf{replacing the center region with input sub-images surrounded by the extrapolated parts}.}
\label{tab:blend}
\end{table}

The comparative methods are all based on image-to-image translation, which need to reconstruct the input sub-image, whilst the sequence-to-sequence based method QueryOTR does not need to reconstruct the input sub-image only outputting the extrapolated regions. In the main manuscript, we report the best results of comparative methods by keeping the reconstructed regions, which were consistent with their original settings. All things being equal, \autoref{tab:blend} reports the results following our setting that replaces the center region with the input sub-image. We additionally report the lower bound of each metric by filling the extrapolated regions with zero pixel values. As shown in \autoref{tab:blend}, our proposed QueryOTR outperforms other methods in most cases, indicating that the higher performance of our method contributes little on the use of the input sub-image. Instead, the results demonstrate the superiority of our method on generating the extrapolated regions. On the other hand, all the comparative methods have an improvement of IS and PSNR metrics due to replacing the center regions with the input sub-image. 

\section{Details of Datasets}
\noindent\textbf{Scenery} is a natural scenery dataset consisting of about 5,000 images in the training set and 1,000 images in the testing set. The images are very diverse and complicated, which contains natural scenes, e.g., snow, valley, seaside, riverbank, sky, and mountain.

\noindent\textbf{Building Facades} is a city scenes dataset consisting of about 16,000 and 1,500 images for training set and testing set respectively. It contains building architecture and city scenes.

\noindent\textbf{WikiArt} is a fine-art paintings dataset obtained from the wikiart.org website. We use the split manner of genres datasets, which contains 45,503 training images and 19,492 testing images.

\section{Inference Time}
The comparison of inference time can be referred in \autoref{tab:time}. Due to the simple but effective design of QueryOTR, our framework is almost three times faster than Uformer which is also engaged with a vision transformer (Swin Transformer) architecture. 

\begin{table}[!h]
\setlength\tabcolsep{7pt}
\centering
\begin{tabular}{c|ccccc}
\hline
Method         & SRN    & NSIPO  & IOH   & Uformer & QueryOTR \\ \hline
Time usage~(ms/image) & 11.960 & 44.190 & 4.160 & 46.810  & 13.345   \\ \hline
\end{tabular}
\caption{Comparison on inference time.}
\label{tab:time}
\end{table}

\section{Hard Examples}
\begin{figure}[b]
\centering
\includegraphics[width=0.96\columnwidth]{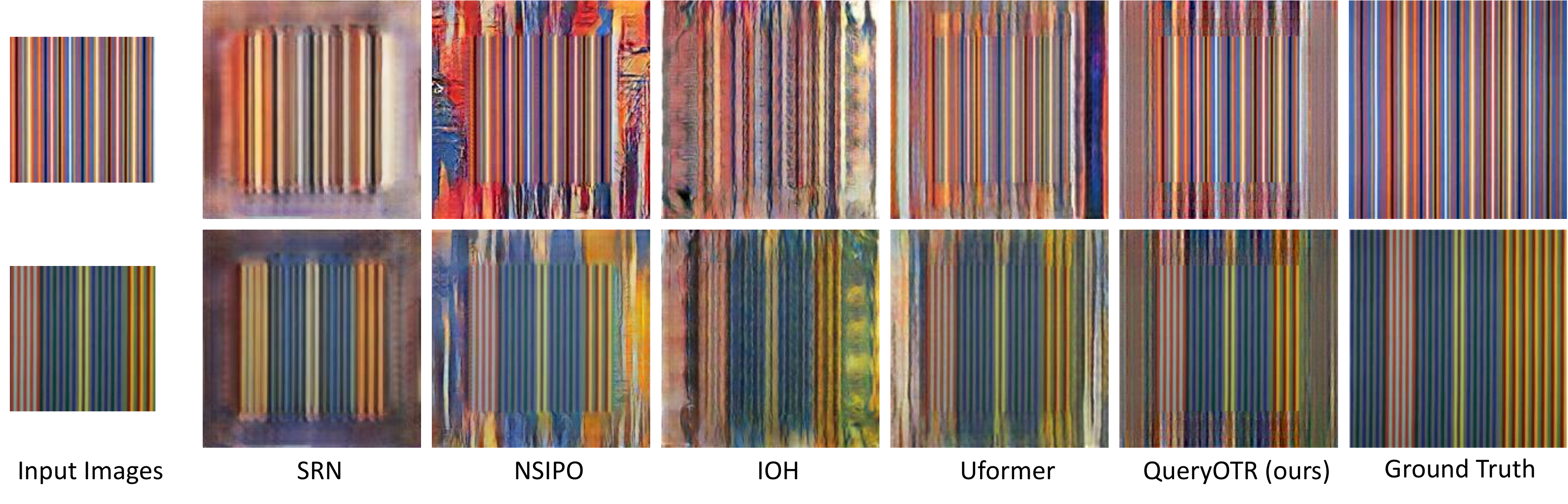}
\caption{Visualization of some hard examples in the test set of WikiArt dataset.}
\label{fig:hardexample}
\end{figure}
We illustrate some hard examples that QueryOTR can work significantly better than the other methods. As shown in \autoref{fig:hardexample}, extrapolating the images with simple colour stripes is very challenging, which requires the network to recognize the pattern and mimic it, especially when such samples are not enough in the training set. The CNN-based methods have limitations to generalize well on such samples, whilst the transformer-based Uformer can generate colorful lines but not straight. In contrast, QueryOTR takes advantage of querying the input sub-image to generate color and straight lines, generating much better images.

\section{More Qualitative Results}
\begin{figure}[h]
\centering
\includegraphics[width=1\columnwidth]{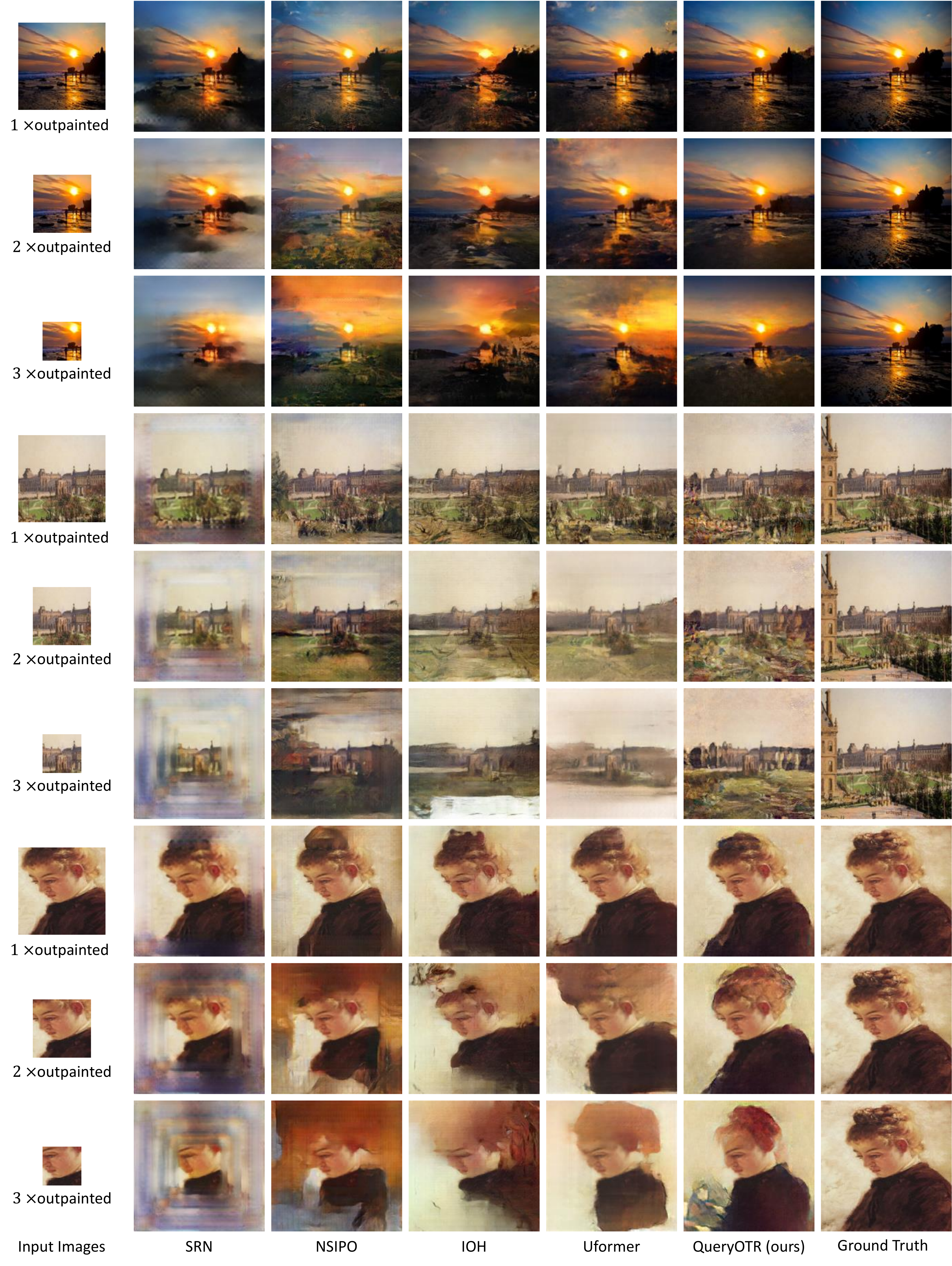}
\caption{Qualitative comparison results on Scenery and WikiArt datasets.}
\label{fig:1}
\end{figure}

\begin{figure}[h]
\centering
\includegraphics[width=1\columnwidth]{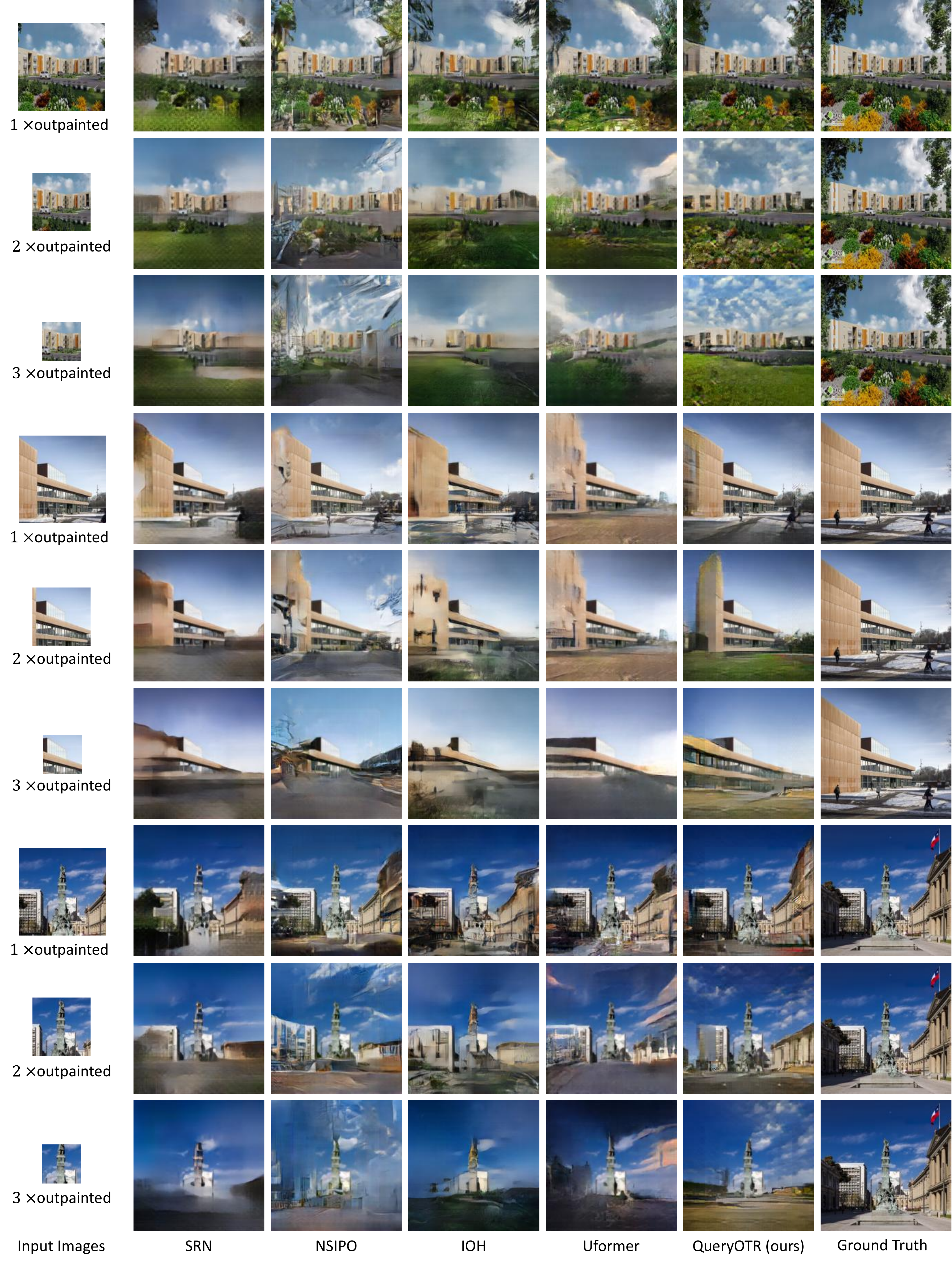}
\caption{Qualitative comparison results on Building Facades dataset.}
\label{fig:2}
\end{figure}

\begin{figure}[h]
\centering
\includegraphics[width=1\columnwidth]{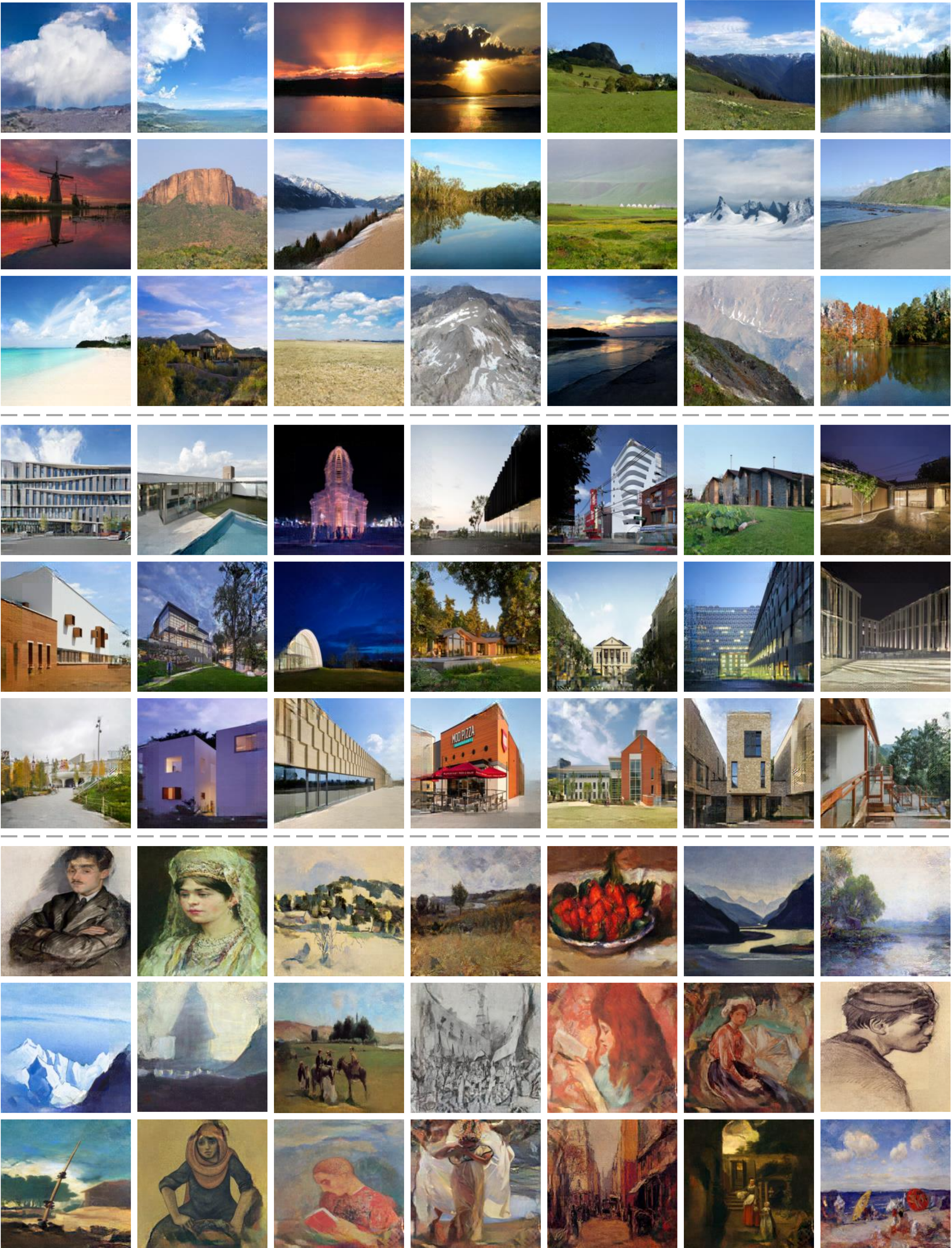}
\caption{Visualization of QueryOTR one-step outpainting.}
\label{fig:3}
\end{figure}

We present more comparative results for one-step and multi-step outpainting. In \autoref{fig:1}, we show additional results on Scenery and WikiArt datasets. Similarly, in \autoref{fig:2}, we provide more results on Building Facades dataset compared with other methods. Meanwhile, we visualize the results conducted by QueryOTR in \autoref{fig:3}.

\end{document}